\pdfoutput=1
\documentclass[letterpaper]{article} % DO NOT CHANGE THIS, use aaai25.sty 
\usepackage{aaai25}  % DO NOT CHANGE THIS
\usepackage{times}  % DO NOT CHANGE THIS
\usepackage{helvet}  % DO NOT CHANGE THIS
\usepackage{courier}  % DO NOT CHANGE THIS
\usepackage[hyphens]{url}  % DO NOT CHANGE THIS
\usepackage{graphicx} % DO NOT CHANGE THIS
\usepackage{natbib}  % DO NOT CHANGE THIS AND DO NOT ADD ANY OPTIONS TO IT
\usepackage{caption} % DO NOT CHANGE THIS AND DO NOT ADD ANY OPTIONS TO IT
\usepackage{algorithm}
\usepackage{algorithmic}

\usepackage{svg}
\usepackage{afterpage}
\usepackage{mathrsfs} % For \mathscr{}
\usepackage{amsmath}  % 提供數學環境
\usepackage{amssymb}  % 提供數學符號，如 \mathbb
\usepackage{multirow} % 引入 multirow 宏包
\usepackage{tabularx} % 引入 tabularx 宏包
\usepackage{booktabs} % 引入 booktabs 宏包

\usepackage{newfloat}
\usepackage{listings}
\DeclareCaptionStyle{ruled}{labelfont=normalfont,labelsep=colon,strut=off} % DO NOT CHANGE THIS
\floatstyle{ruled}
\newfloat{listing}{tb}{lst}{}
\floatname{listing}{Listing}

\title{KeyGS: A Keyframe-Centric Gaussian Splatting Method for Monocular Image Sequences}
\author{
   Keng-Wei Chang,
   Zi-Ming Wang, 
   Shang-Hong Lai
}
\affiliations{
   National Tsing Hua University\\
   percyx987654321@gapp.nthu.edu.tw, ziming614@gapp.nthu.edu.tw, lai@cs.nthu.edu.tw
}

\begin{document}
\maketitle

% Figure1
\begin{figure*}[t]
    \centering
    % \captionsetup{type=figure, aboveskip=5pt}
    \includegraphics[width=1.7\columnwidth]{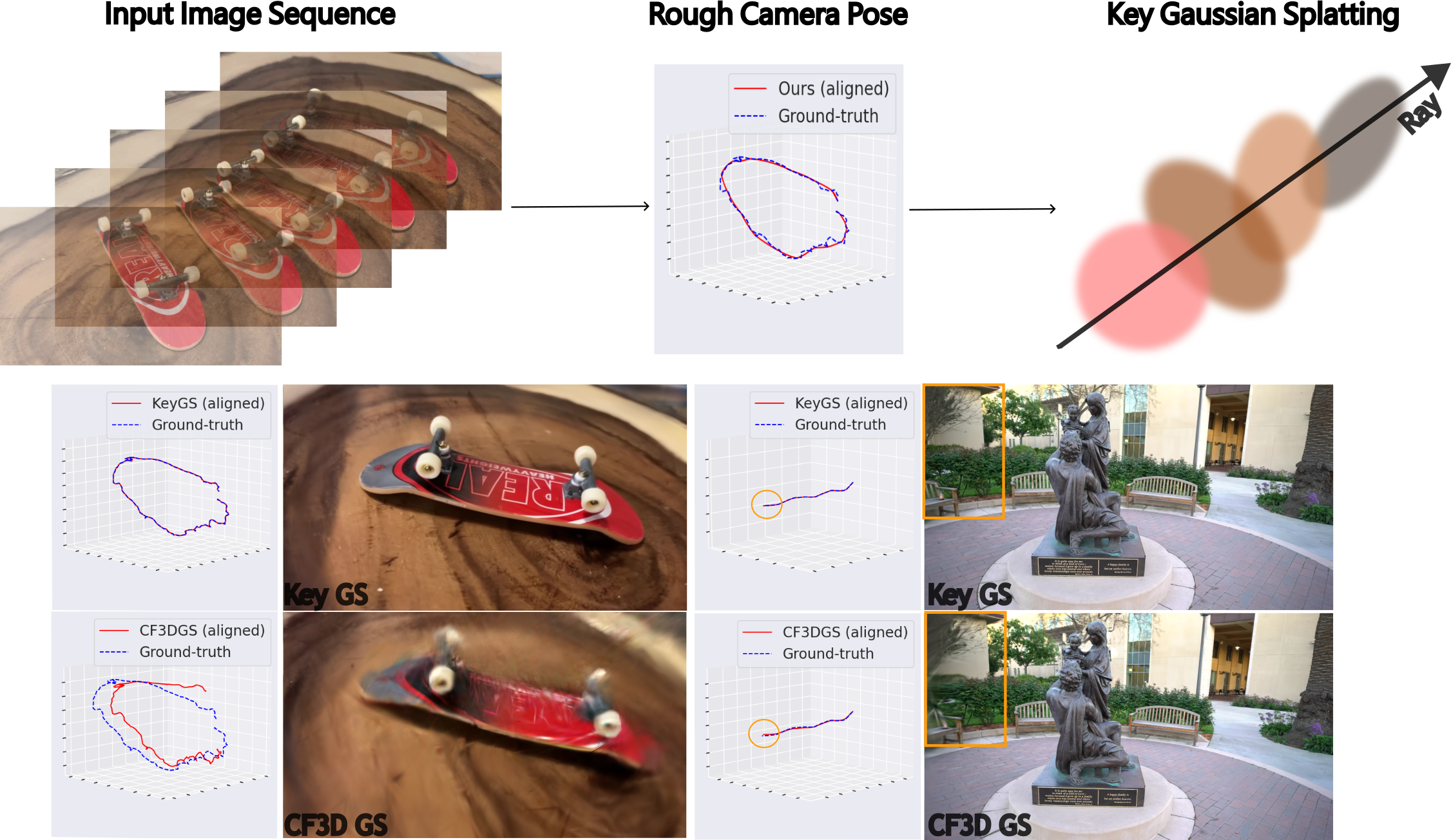}
    \caption{
        \textbf{KeyGS Framework}. For each sequence of images, we sub-sample {$\frac{1}{N}$} images as keyframes and perform fast, albeit less accurate, sequential \textbf{SFM} with second to obtain an initial rough trajectory. We then jointly optimize the camera poses using the \textbf{KeyGS} method. Compared to \textbf{CF3DGS}, \textbf{KeyGS} continuously refines the camera poses to reduce accumulation errors that lead to localization drift. Additionally, \textbf{KeyGS} achieves detailed reconstruction by refining camera poses. }
    \label{fig:teaser}
\end{figure*}

\begin{abstract}
\setlength{\parindent}{1em}
\begin{em}
Reconstructing high-quality 3D models from sparse 2D images has garnered significant attention in computer vision. Recently, \textbf{3D Gaussian Splatting (3DGS)} has gained prominence due to its explicit representation with efficient training speed and real-time rendering capabilities. However, existing methods still heavily depend on accurate camera poses for reconstruction. Although some recent approaches attempt to train \textbf{3DGS} models without the \textbf{Structure-from-Motion (SfM)} preprocessing from monocular video datasets, these methods suffer from prolonged training times, making them impractical for many applications.

In this paper, we present an efficient framework that operates \textbf{without any depth or matching model}. Our approach initially uses \textbf{SfM} to quickly obtain rough camera poses within seconds, and then refines these poses by leveraging the dense representation in \textbf{3DGS}. This framework effectively addresses the issue of long training times. Additionally, we integrate the densification process with joint refinement and propose a \textbf{coarse-to-fine frequency-aware densification} to reconstruct different levels of details. This approach prevents camera pose estimation from being trapped in local minima or drifting due to high-frequency signals. Our method significantly reduces training time from hours to minutes while achieving more accurate novel view synthesis and camera pose estimation compared to previous methods.
\end{em}
\end{abstract}


\section{Introduction}
% introduce 3D reconstruction
In recent years, 3D photorealistic reconstruction has gained popularity, especially with differential rendering techniques \cite{3DGS,NeRF,NeGF,NeuS,PointNeRF}. These methods use a novel approach, representing the 3D model as a differentiable volume field or a traditional representation, and optimize it through a differential rendering pipeline, leading to exceptionally high-quality reconstructions.

% introduce NeRF and 3DGS
Notable representations include \textbf{Neural Radiance Fields} (\textbf{NeRF}) \cite{NeRF} and the recently popular \textbf{3D Gaussian Splatting} (\textbf{3DGS}) \cite{3DGS}. Both methods use volume rendering \cite{VRDigest}, but they differ significantly in their approaches. \textbf{NeRF} employs ray-marching, which leads to slow inference due to the high computational demands of sampling along rays and feeding data to an MLP. In contrast, \textbf{3DGS} uses differential rasterization without an MLP, enabling real-time inference speeds. 

% Introduce the of data preprocess SfM
In many \textbf{NeRF} and \textbf{3DGS} reconstruction pipelines, a common approach involves using software like \textbf{COLMAP} \cite{COLMAP} for \textbf{Structure from Motion} (\textbf{SfM}) to estimate camera poses. \textbf{SfM} extracts SIFT features from images, applies the \textbf{RANSAC} algorithm for pose estimation, and performs bundle adjustment for refinement. However, this method often struggles under extreme conditions, such as noisy images, textureless regions, low resolution, or varying lighting, leading to inaccurate pose estimates. Moreover, \textbf{SfM} becomes computationally expensive with an increasing number of images due to the complexity of \textbf{RANSAC} and pairwise bundle adjustment. To overcome these challenges, some researchers focus on refining camera poses during 3D reconstruction or on methods that avoid \textbf{SfM} altogether.

% introduce the work with joint refinement
To further refine camera pose estimates in \textbf{SfM}, various approaches\cite{BARF,JTensoRF,SCNeRF,CF3DGS,BAA,RoNGP,CamP,L2GNERF,GNeRF,LocalRF,NoPe,NeRFmm,ParallelNeRF,iMAP,GSSLAM,NiCESLAM} have been proposed to jointly refine camera poses, either starting from noisy initial poses or without any initial pose information. \textbf{Bundle-Adjusting Neural Radiance Fields (BARF)}\cite{BARF} is the first \textbf{NeRF} method to use dense photometric signals for alignment. It also introduces a heuristic \textbf{coarse-to-fine strategy} that progressively increases the signal frequency to effectively refine camera poses.
\textbf{Joint TensoRF}\cite{JTensoRF} provides theoretical analysis indicating that image alignment may encounter gradient oscillation with high-frequency signals. To address this issue, it employs Gaussian filter to reduce frequency and utilizes a grid-based \textbf{NeRF}, \textbf{TensoRF} \cite{TensoRF} as their representation. There are also more extreme approaches that completely avoid using any initial camera pose. \textbf{Nope-NeRF}\cite{NoPe} considers neighboring sequences and uses a monocular depth estimation model like \textbf{DPT} \cite{DPT} to minimize reprojection error. Recently, with the emergence of \textbf{3DGS}, \textbf{COLMAP-Free 3D Gaussian Splatting} \textbf{(CF3DGS)}\cite{CF3DGS} also uses the \textbf{DPT} to predict depth map as a prior to progressively register camera poses, achieving excellent performance.
% roughly introduce our method 

However, the methods mentioned above suffer from long training times, ranging from hours to days, and NeRF-based approaches are constrained by the ray-marching rendering pipeline. \textbf{CF3DGS} employs progressive registration of camera poses and stops refining them after registration, which can lead to inefficient training speeds and accumulated errors, as illustrated by the trajectory in Figure \ref{fig:teaser}.

To address these issues, we propose \textbf{KeyGS}, which uses a fast but less accurate \textbf{SfM} method to quickly obtain rough camera poses in seconds. \textbf{KeyGS} then jointly refines both pose and reconstruction, enabling complete training in approximately 10 minutes. Moreover, our approach continuously refines the camera poses to alleviate the error accumulation problem.
In summary, this paper makes the following contributions:

\begin{itemize}
\item We propose a framework that combines \textbf{Structure-from-Motion} with \textbf{3DGS} to efficiently obtain initial rough camera poses and then refine them using \textbf{3DGS}, addressing the limitations of traditional pipelines.
\item We introduce a joint refinement approach that continuously improves camera poses, mitigating error accumulation and enhancing reconstruction accuracy.
\item We develop a \textbf{coarse-to-fine frequency-aware densification} technique, which builds a relationship between signal alignment and densification to refine camera poses and reduce artifacts in the reconstruction process.
\end{itemize}

% Figure2
\begin{figure*}[t]
    \centering
    \includegraphics[width=1.6\columnwidth]{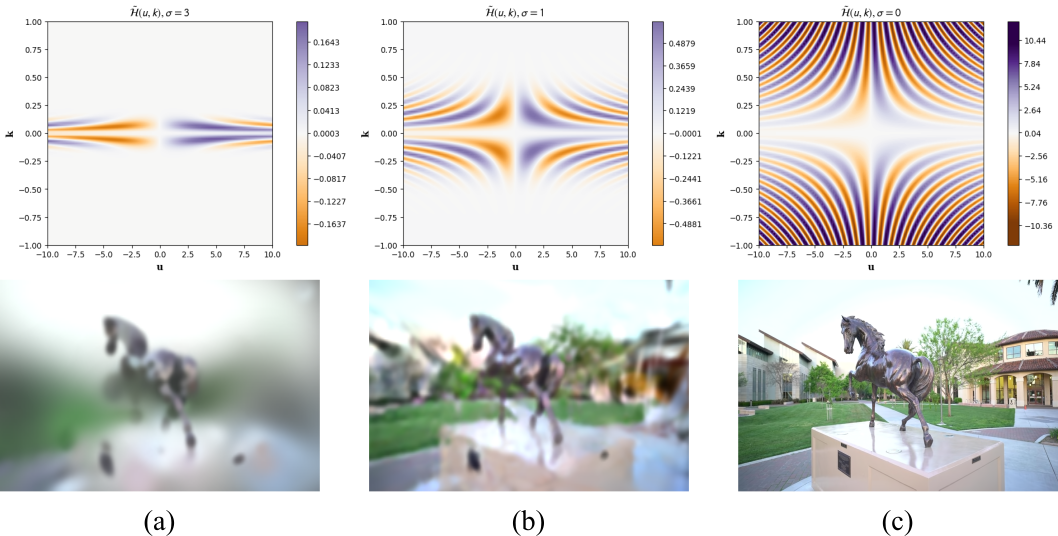}
    % \captionsetup{type=figure, aboveskip=5pt}
    \caption{
        \textbf{Illustration of Coarse-to-Fine Frequency-Aware Densification –} The top part shows the gradient for each frequency related to alignment offset, using different scales $\sigma$ for Gaussian smoothing on the Fourier kernel $\Tilde{\mathcal{H}}(u,k)$. Larger $\sigma$ values concentrate the gradient on low frequencies, while decreasing $\sigma$ shifts it to higher frequencies. The bottom part visualizes the training process for various $\sigma$ values. At high $\sigma$ \textbf{(a)}, there are no details. As $\sigma$ decreases, rough contours emerge \textbf{(b)}, and densification is primarily influenced by high-frequency gradients, leading to detailed structures at low $\sigma$ \textbf{(c)}.}   
    \label{fig:method}
\end{figure*}

\section{Related Work}
\subsection{Novel view synthesis}
Recent advancements in novel view synthesis have been notably propelled by the development of differential rendering. \textbf{NeRF} uses an \textbf{MLP} to model both the geometry and view-dependent appearance of scenes. By optimizing through ray marching with \textbf{volume rendering} \cite{VRDigest} techniques, \textbf{NeRF} achieves impressive rendering quality.

However, \textbf{NeRF} can be very inefficient because it requires a large number of sample points to be fed into a deep \textbf{MLP}, using shared weights to represent the entire scene. To improve efficiency, some works \cite{iNGP,TensoRF,planoctree,DVGO} utilize spatial data structures, such as grids or octrees, to optimize specific learnable features stored in these structures within 3D space. Despite these enhancements, the computationally expensive process of ray marching remains a limitation for wider applications, as it prevents real-time rendering.

To overcome these challenges, \textbf{3DGS} introduces a differential rasterization approach that represents 3D models as a set of Gaussian spheres and uses \textbf{volume rendering} to blend colors projected from these 3D Gaussians. This shift in the rendering pipeline results in higher inference speeds and a more explicit representation, garnering significant attention across various research domains.

% SFM and the time cost analysis 
\subsection{Structure From Motion}
Even methods like \textbf{NeRF} and \textbf{3DGS} typically require preprocessing to obtain accurate camera poses, as these reconstruction techniques are highly sensitive to camera pose precision. The most commonly used algorithm for camera pose estimation is \textbf{SfM}, with \textbf{COLMAP} being one of the most well-known and widely used software tools. It employs an incremental \textbf{SfM} method involving three main steps: feature extraction, feature matching with \textbf{RANSAC}, and bundle adjustment. Since it uses exhaustive mode to match and perform bundle adjustment on all pairs of images, its computational complexity can be quite high, approximately $O(n^4)$.

However, \textbf{COLMAP} offers a sequential matching mode for image sequences, which matches $M$ neighbors and performs bundle adjustment among them, this mode is often less accurate for camera pose estimation. As a result, most tasks do not rely on this mode for registering camera poses with sequential data. In contrast,  methods like \textbf{Hierarchy GS} \cite{HierarchyGS} are designed to work at large scales, utilizing sequential data while still opting for the exhaustive mode to obtain more accurate camera poses.

% joint Refine camera pose with NeRF and 3DGS (point cloud initialization , training time)
\subsection{Joint Refinement}
Although \textbf{COLMAP} can provide nearly accurate camera poses, it can still fail under certain conditions such as noisy images, a limited number of images, or reflective surfaces. To address these issues, several models have been developed to jointly refine camera poses and improve 3D reconstruction accuracy.

\textbf{BARF} \cite{BARF} is among the pioneering models that establish a link between camera pose estimation and 3D reconstruction. While \textbf{NeRFmm} \cite{NeRFmm} refines intrinsic and extrinsic parameters using camera pose embeddings, \textbf{NoPeNeRF} \cite{NoPe} reconstructs image sequences by leveraging a depth estimator \textbf{DPT} \cite{DPT} to obtain pseudo-depth information and jointly refines camera poses with reprojection error from neighboring images as regularization. 

Similarly, \textbf{CF3DGS} \cite{CF3DGS} also utilizes a depth estimator \textbf{DPT} to initialize the 3D point cloud, which is crucial step in \textbf{3DGS}. However, \textbf{CF3DGS}\cite{CF3DGS} fixes the camera poses once they are registered, which can lead to accumulated errors in pose estimation, as shown in Figure \ref{fig:teaser}. Additionally, we observe that the error accumulation problem can cause \textbf{CF3DGS}\cite{CF3DGS} to excessively split the Gaussian spheres, resulting in increased memory usage and potential crashes during training due to memory limitations.

\subsection{Coarse-to-Fine Strategy}
\textbf{BARF} \cite{BARF} found that naive joint refinement fails to recover accurate poses due to difficulties in alignment caused by high-frequency components. Instead, they employ a coarse-to-fine strategy to progressively reveal the frequency components of positional encoding in vanilla \textbf{NeRF} \cite{NeRF}, which enables effective joint refinement of camera poses and scene reconstruction.

Although \textbf{BARF} \cite{BARF} experimentally validates the effectiveness of the coarse-to-fine strategy, it is limited to vanilla \textbf{NeRF} \cite{NeRF}. \textbf{Joint TensoRF} \cite{JTensoRF} is a \textbf{NeRF} model that focuses on joint refinement by utilizing grid-based \textbf{NeRF} \cite{TensoRF} for reconstruction. Previous works \cite{RoNGP,BAA} have shown that grid features can be unsuitable for joint refinement of camera poses due to discretization and gradient oscillations. However, \textbf{Joint TensoRF} \cite{JTensoRF} addresses this issue by providing general theoretical analysis and solution of joint refinement. They use a 1D signal \( f_{gt} \), as an example. To align two 1D signals with a shift offset \( u \), the gradient related to offset \( u \) is given by Equation~\ref{eq:fourier}:
\begin{equation}
    \footnotesize
    \frac{d}{du}\mathcal{L}=\int \| \mathcal{F}[f_{gt}]\|^2 \mathcal{H}(u,k) \, dk
    \label{eq:fourier}
\end{equation}
where $\mathcal{H}(u,k) = 4 \pi k \sin(2 \pi k u)$, $\mathcal{F}[f_{gt}]$ is the Fourier transform of $f_{gt}$, and $k$ is the wavenumber in the frequency domain. Intuitively, the kernel $\mathcal{H}(u,k)$ transforms the spectrum $\mathcal{F}[f_{gt}]$ into the derivative $\frac{d}{du}\mathcal{L}$. The kernel $\mathcal{H}(u,k)$ exhibits gradient oscillations at high frequencies and is quasi-convex at low frequencies. To address the challenges posed by high frequencies, the authors apply \textbf{coarse-to-fine} Gaussian smoothing to the signal. Consequently, the transform kernel $\mathcal{H}(u,k)$ becomes $\tilde{\mathcal{H}}(u,k) = \| \mathcal{F}[\mathcal{N}(0,\sigma^2)]\|^2 \cdot \mathcal{H}(u,k)$, which enhances the effectiveness of joint refinement in voxel-based NeRF, \textbf{TensoRF}. For a visual representation, we refer the reader to Figure~\ref{fig:method}.
In summary, while existing methods aim to refine camera poses and use sequential image structures, they often suffer from long training times due to inefficient pipelines. This limits their application for real-world uses. Our work overcomes these challenges with the proposed \textbf{KeyGS}, an efficient framework leveraging \textbf{3DGS}. We also develop a \textbf{coarse-to-fine frequency-aware densification} strategy, inspired by \textbf{Joint TensorRF} \cite{JTensoRF}, for effective and practical camera pose refinement.

\section{Proposed Method}
Given a sequence of images $\{I_i^{gt}\}_{i=1}^{N}$, our goal is to efficiently recover camera poses $\{T_i\}_{i=1}^{N}$ and generate a photorealistic 3D scene, simultaneously. We propose \textbf{KeyGS}, a framework designed to quickly refine initial rough camera poses. It then jointly reconstructs the scene and refines the camera poses using \textbf{3DGS} along with our coarse-to-fine frequency-aware densification technique.

\subsection{Preliminary: 3D Gaussian Splatting}
% 簡單講一下3DGS 應該不用提及 NeRF 並且需要講到他需要 attribute
\textbf{3DGS} \cite{3DGS} represents a scene using explicit 3D Gaussians, unlike \textbf{NeRF}'s implicit representation. It uses rasterization to project the 3D Gaussians onto the image plane and applies volume rendering to blend colors. This approach is efficient for inference due to its active projection and the absence of MLP involvement.
\begin{equation}
    \footnotesize
    G(x)=e^{-\frac{1}{2}(x-\mu)^T\Sigma^{-1}(x-\mu)}
\label{eq:gaussian}
\end{equation}
% 借由initial point  去 inrtroduce 3dgs 需要優化的參數
However, \textbf{3DGS} requires appropriate initial conditions, often utilizing a point cloud estimated by SfM with position $ \mu \in \mathbb{R}^3 $, color $ c \in \mathbb{R}^3 $ (parameterized by spherical harmonics), and initialized with low opacity $ o \in \mathbb{R}^1 $. In order to represent the 3D Gaussians in Equation \ref{eq:gaussian}, it employs \textbf{KNN} for estimating the covariance matrix $\Sigma \in \mathbb{R}^{3 \times 3}$. During optimization, $\Sigma$ is decomposed into a rotation $ R \in \mathbb{R}^{3 \times 3} $ and a scale $ S \in \mathbb{R}^{3 \times 3} $ to ensure it remains positive semi-definite, as shown in Equation \ref{eq:Sigma}.

\begin{equation}
    \footnotesize
    \Sigma=RSS^TR^T
    \label{eq:Sigma}
\end{equation}
% Figure3
\begin{figure}[t]
    \centering
    \includegraphics[width=0.8\columnwidth]{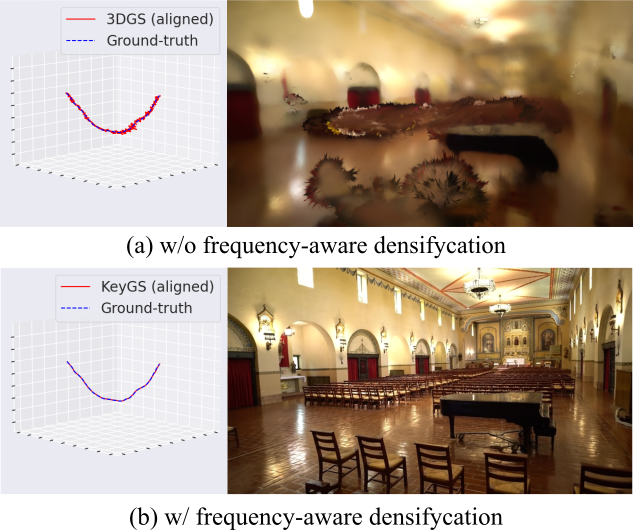}
    % \captionsetup{type=figure, aboveskip=5pt}
    \caption{
        \textbf{Comparison of naive joint refinement and our proposed method, frequency-aware densification.} (a) Applying naive joint refinement to \textbf{3DGS} results in over-splitting of the Gaussians to fit high-frequency signals. This causes the Gaussians to become spiky, making alignment more difficult and leading to oscillations in the trajectory. (b) Our proposed frequency-aware densification method uses a coarse-to-fine approach to account for gradients of different frequencies. The reconstruction results are smoother and more accurate, leading to improved camera pose recovery.}
    \label{fig:densification}
\end{figure}
To represent the scene using 3D Gaussians, rasterization is employed to render them as images. Given the camera view $W \in SE(3)$ and camera intrinsic $K \in \mathbb{R}^{3 \times 4}$, the 3D Gaussians are projected onto the image plane. This projection involves using a Taylor expansion to approximate the 3D Gaussians as 2D Gaussians on the plane. The projected mean $\mu^{2D} \in \mathbb{R}^{2}$ is computed by applying the camera view $W$ and the camera intrinsic matrix $K$ to $\mu$. Meanwhile, the covariance $\Sigma^{2D} \in \mathbb{R}^{2 \times 2}$ is determined by using the Jacobian $J$ of the camera projection, as shown in Equation \ref{eq:Jabcobian}.
\begin{equation}
    \footnotesize
    \Sigma^{2D}=JW\Sigma W^TJ^T
    \label{eq:Jabcobian}
\end{equation}
Finally, using volume rendering as described in Equation \ref{eq:Volrender}, the colors $c_i$ of each Gaussian are blended sequentially based on their depth, with the alpha value $\alpha_i(x) = o_i(x)G_i(x)$ associated with each Gaussian. This process results in the rendered output $\hat{C}(x)$, which is then compared to the ground truth $C^{gt}(x)$ to compute the loss.

\begin{equation}
    \footnotesize
    \hat{C}(x)=\sum\limits_{i=1}^{N} c_i \alpha _i(x) \prod\limits_{j=1}^{i-1} (1-\alpha_j(x))
    \label{eq:Volrender}
\end{equation}

\subsection{KeyGS Framework}
To address the efficiency issues related to the time cost of feature matching and bundle adjustment in the \textbf{SfM} problem, we use the sequential mode of \textbf{COLMAP} instead of the exhaustive mode to match features within a small neighborhood of images. This approach reduces the time complexity from $O(n^4)$ to $O(n^2)$, significantly lowering the computational cost of \textbf{SfM}. Furthermore, since the cost is highly dependent on the number of images, we leverage the sequential structure of the image set by uniformly subsampling $\frac{1}{N}$ of the images in the sequence and applying sequential \textbf{SfM} to these keyframes. After applying \textbf{SfM}, we obtain the camera poses represented by quaternions $q_i \in \mathbb{R}^4$ and translations. For the remaining unsampled images, we perform \textbf{spherical linear interpolation (SLERP)} for quaternions (Equation \ref{eq:slerp}) and linear interpolation for translations. This framework efficiently estimates rough camera poses in seconds for the trajectory, as illustrated in Figure \ref{fig:teaser}.

\begin{equation}
    \footnotesize
    Slerp(q_i,q_{i+1},t)=\frac{sin(1-t)\theta }{sin(\theta)}q_i + \frac{sin(t\theta)}{sin(\theta)} q_{i+1}
    \label{eq:slerp}
\end{equation}
Once the rough trajectory is obtained, we employ the \textbf{KeyGS} with our \textbf{frequency-aware densification} to the joint refinement process, utilizing dense RGB information to further refine the camera poses. This step ensures that the initial estimations are improved upon, allowing for a more accurate reconstruction of the scene.

\subsection{Joint Refinement and Densification}
% problem formula 
To refine the camera pose $T_i \in SE(3)$ for each image $I_i$, we train \textbf{3DGS} with a learnable $SE(3)$ transformation $\Delta T_i$ to obtain the estimated pose $T_{i}^{\text{pred}}=\Delta T_i T_i$, Specifically, we do not parameterize $\Delta T_i$ using $\mathfrak{se}(3)$; instead, we use $\mathfrak{so}(3) \times \mathfrak{t}(3)$ to reduce the influence between rotation and translation.
The image is rendered using rasterization from 3DGS, which depends on the predicted camera pose $T^{\text{pred}}_{i}$ and the optimized Gaussians $G$. We minimize the photometric loss function $\mathcal{L}_{\text{rgb}}$, which integrates $\mathcal{L}_1$ and D-SSIM losses , balanced by a factor $\lambda$, as detailed in Equation \ref{eq:rgb_loss}. The attributes of the Gaussians and the pose refinements $\Delta T_i$ are updated through backpropagation.
\begin{equation}
    \footnotesize
    \mathcal{L}_{rgb}=(1-\lambda)\mathcal{L}_1+\lambda \mathcal{L}_{\text{D-SSIM}}
    \label{eq:rgb_loss}
\end{equation}
% densification
Another important factor in joint refinement is the densification process. In \textbf{3DGS}, Gaussians are split if the gradient with respect to their position $\mu^{2D}$, $\frac{\partial L}{\partial \mu^{2D}}$, exceeds a certain threshold, and they are culled if their opacity, $o$, falls below a specified threshold. Both the refinement and densification processes are primarily influenced by the gradient term $\frac{\partial \mathcal{L}}{\partial G} \frac{\partial G}{\partial \mu^{2D}}$, as described in Equation \ref{eq:gradient}. Notably, only the 2D splat affects these processes.
\begin{equation}
    \footnotesize
    \frac{\partial L}{\partial \Delta T}=\frac{\partial L}{ \partial G} \frac{\partial G}{ \partial \mu^{2D}}\frac{\partial \mu^{2D}}{ \partial \Delta T}
    \label{eq:gradient}
\end{equation}
As shown in Figure \ref{fig:method} (c), high-frequency signals can lead to misplaced Gaussians, causing excessive splitting due to larger, oscillating gradients. Excessive splitting results in more misaligned Gaussians and an increases probability of trapping in local minima due to overfitting with incorrect poses, which complicate the optimization process and creating a vicious cycle. We depict an example for visualization in Figure \ref{fig:densification}.

% Figure4
\begin{figure}[h]
    \centering
    \includegraphics[width=0.9\columnwidth]{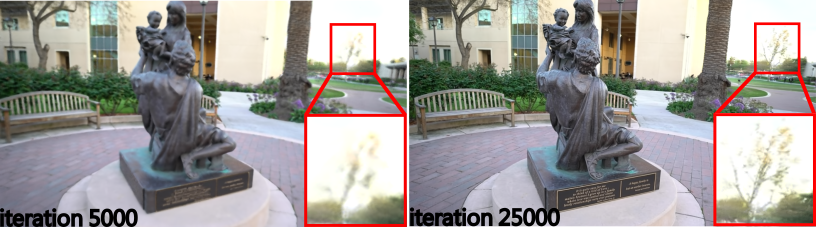}
    % \captionsetup{type=figure, aboveskip=5pt}
    \caption{
        \textbf{Coarse-to-Fine Frequency-Aware Densification}. Our method aligns signals by preventing premature Gaussian splitting at high frequencies. Although the signal may appear aligned early on, this approach suppresses gradient influence from high-frequency details. As the Gaussian filter scale decreases, the gradient shifts to high frequencies, enabling Gaussians to split and capture finer details.}
    \label{fig:densification_detail}
\end{figure}

\subsection{Coarse To Fine Frequency-Aware Densificaion}
To address gradient oscillation and excessive splitting issues, we propose a \textbf{coarse-to-fine frequency-aware} densification. First, we visualize the kernel $\tilde{\mathcal{H}}(u,k)$, as mentioned in related work, with varying scales $\sigma$ of Gaussian blur in Figure \ref{fig:method}. We observe that Gaussian smoothing not only suppresses high-frequency gradients but also adjusts the dominant gradient to different frequency levels as the scale $\sigma$ changes. This approach is particularly effective for densification with misaligned Gaussians, as it provides stable gradients for low frequencies, preventing excessive splitting and enabling effective signal alignment.
\begin{equation}
    \footnotesize
    \begin{aligned}
        \frac{d}{du}\mathcal{L} &= \int \| \mathcal{F}[f_{gt}] \|^2 \tilde{\mathcal{H}}(u,k) \, dk \\
        &= \int \| \sum\limits_{i}^{N} a_i \mathcal{F}[ \mathcal{N}(0,\sigma^2)\ast G_i] \|^2 \mathcal{H}(u,k) \, dk
    \end{aligned}
    \label{eq:gmm_gradient}
\end{equation}
We analyze how Gaussian blur affects the gradient with 3DGS, as described in Equation \ref{eq:gmm_gradient}. Since the signal is represented by the blending of Gaussians using Equation \ref{eq:Volrender}, it can be viewed as linear combination of Gaussians. By leveraging the linear property of the Fourier transform, we find that applying a smooth gradient operator is equivalent to applying Gaussian blur to each Gaussian $G_i$.
\begin{equation}
    \footnotesize
    G_i(x, \sigma) = \sqrt{\frac{\left|\Sigma\right|}{\left|\Sigma+\sigma^2 I\right|}}e^{-\frac{1}{2}(x - \mu)^T (\Sigma + \sigma^2 I)^{-1} (x - \mu)}
    \label{eq:gaussian_filter}
\end{equation}
Therefore, we apply a Gaussian filter to each Gaussian $G_i$ as defined in Equation \ref{eq:gaussian_filter} . Initially, a large-scale filter suppresses gradient influence from high-frequency details. As training progresses, we gradually reduce the filter scale $\sigma$, enhancing the ability to capture fine details. This ensures effective signal alignment. as shown in Figure \ref{fig:densification_detail}.
% Table1, Tanks and Temples PSNR
\begin{table*}[t]
    \small
    \centering
    \begin{tabularx}{\textwidth}{lXcccXcccXcccXcccXccc} % 使用 tabularx 环境，并指定列类型
  % \toprule
  \hline
  \multirow{2}{*}{Scenes} &  & \multicolumn{3}{c}{KeyGS (ours)} &  & \multicolumn{3}{c}{CF3DGS} &  & \multicolumn{3}{c}{Nope-NeRF} &  & \multicolumn{3}{c}{BARF} &  & \multicolumn{3}{c}{SC-NeRF} \\
  \cline{3-5} \cline{7-9} \cline{11-13} \cline{15-17} \cline{19-21} &  & PSNR$\uparrow$ & SSIM$\uparrow$ & LPIPS$\downarrow$ &  & PSNR & SSIM  & LPIPS &  & PSNR & SSIM & LPIPS &  & PSNR & SSIM & LPIPS &  & PSNR & SSIM & LPIPS \\ 
  \hline
  Church &  & \textbf{30.62} & \textbf{0.92} & \textbf{0.06} &  & 30.23 & 0.93 & 0.11 & & 25.17 & 0.73 & 0.39 &  & 23.17 & 0.62  & 0.52 &  & 21.96 & 0.60 & 0.53\\
  Barn &  & \textbf{34.25} & \textbf{0.95} & \textbf{0.04} &  & 31.23 & 0.90 & 0.10 & & 26.35 & 0.69 & 0.44 &  & 25.28 & 0.64 & 0.48 &  & 23.26 & 0.62 & 0.51\\
  Museum &  & \textbf{33.46} & \textbf{0.94} & \textbf{0.03} &  & 29.91 & 0.91 & 0.11 & & 26.77 & 0.76 & 0.35 &  & 23.58 & 0.61 & 0.55 &  & 24.94 & 0.69 & 0.45\\
  Family &  & \textbf{33.05} & \textbf{0.95} & \textbf{0.04} &  & 31.27 & 0.94 & 0.07 & & 26.01 & 0.74 & 0.41 &  & 23.04 & 0.61 & 0.56 &  & 22.60 & 0.63 & 0.51\\
  Horse &  & 33.65 & \textbf{0.96} & \textbf{0.03} &  & \textbf{33.94} & \textbf{0.96} & 0.05 & & 27.64 & 0.84 & 0.26 &  & 24.09 & 0.72 & 0.41 &  & 25.23 & 0.76 & 0.37\\
  Ballroom &  & \textbf{33.70} & 0.95 & \textbf{0.02} &  & 32.47 & \textbf{0.96} & 0.07 & & 25.33 & 0.72 & 0.38 &  & 20.66 & 0.50 & 0.60 &  & 22.64 & 0.61 & 0.48\\
  Francis &  & \textbf{34.45} & \textbf{0.93} & \textbf{0.08} &  & 32.72 & 0.91 & 0.14 & & 29.48 & 0.80 & 0.38 &  & 25.85 & 0.69 & 0.57 &  & 26.46 & 0.73 & 0.49\\
  Ignatius &  & \textbf{30.85} & \textbf{0.92} & \textbf{0.06} &  & 28.43 & 0.90 & 0.09 & & 23.96 & 0.61 & 0.47 &  & 21.78 & 0.47 & 0.60 &  & 23.00 & 0.55 & 0.53\\ 
  \hline
  Mean &  & \textbf{33.01} & \textbf{0.94} & \textbf{0.04} &  & 31.28 & 0.93 & 0.09 & & 26.34 & 0.74 & 0.39 &  & 23.42 & 0.61 & 0.54 &  & 23.76 & 0.65 & 0.48\\ 
  % \bottomrule
  \hline
    \end{tabularx}
% \captionsetup{type=table, aboveskip=5pt}
\caption{
    \textbf{Novel view synthesis results on Tanks and Temples.} Each baseline method is trained with its public code under the original settings and evaluated with the same evaluation protocol. The best results are highlighted in bold.}
\label{table:tanks_psnr}
\end{table*}

% Table 2, Tanks and Temples trakectory
\begin{table*}[t]
\small
\centering
\begin{tabularx}{\textwidth}{lXcccXcccXcccXcccXccc}
  % \toprule
  \hline
  \multirow{2}{*}{Scenes} &  & \multicolumn{3}{c}{KeyGS (ours)} &  & \multicolumn{3}{c}{CF3DGS}  &  & \multicolumn{3}{c}{Nope-NeRF} &  & \multicolumn{3}{c}{BARF} &  & \multicolumn{3}{c}{SC-NeRF} \\ 
  \cline{3-5} \cline{7-9} \cline{11-13} \cline{15-17} \cline{19-21} &  & $\text{RPE}_t$$\downarrow$ & $\text{RPE}_r$$\downarrow$ & ATE$\downarrow$ &  & $\text{RPE}_t $ & $\text{RPE}_r $ & ATE &  &  $\text{RPE}_t $ & $\text{RPE}_r $ & ATE &  & $\text{RPE}_t$ & $\text{RPE}_r$ & ATE   &  & $\text{RPE}_t$ & $\text{RPE}_r$ & ATE \\ 
  \hline
   Church &  & \textbf{0.006}& 0.013 & \textbf{0.000}      &  & 0.008 & 0.018 & 0.002      & & 0.034 & \textbf{0.008} & 0.008      &  & 0.114 & 0.038  & 0.052       &  & 0.836 & 0.187 & 0.108\\
   Barn      &  & \textbf{0.008} & \textbf{0.016} & \textbf{0.001}      &  & 0.034 & 0.034 & 0.003      & & 0.046 & 0.032 & 0.004      &  & 0.314 & 0.265 & 0.050      &  & 1.317 & 0.429 & 0.157\\
   Museum      &  & \textbf{0.025} & \textbf{0.025} & \textbf{0.002}      &  & 0.052 & 0.215 & 0.005      & & 0.207 & 0.202 & 0.020      &  & 3.442 & 1.128 & 0.263      &  & 8.339 & 1.491 & 0.316\\
   Family      &  & \textbf{0.012} & \textbf{0.012} & \textbf{0.000}      &  & 0.022 & 0.024 & 0.002      & & 0.047 & 0.015 & 0.001      &  & 1.371 & 0.591 & 0.115      &  & 1.171 & 0.499 & 0.142\\
   Horse       &  & \textbf{0.078} & \textbf{0.002} & \textbf{0.001}      &  & 0.112 & 0.057 & 0.003      & & 0.179 & 0.017 & 0.003      &  & 1.333 & 0.394 & 0.014      &  & 1.336 & 0.438 & 0.019\\
   Ballroom      &  & \textbf{0.015} & \textbf{0.014} & \textbf{0.000}      &  & 0.037 & 0.024 & 0.003      & & 0.041 & 0.018 & 0.002      &  & 0.531 & 0.228 & 0.018      &  & 0.328 & 0.146 & 0.012\\
   Francis      &  & \textbf{0.007} & 0.016 & \textbf{0.001}     &  & 0.029 & 0.154 & 0.006      & & 0.057 & \textbf{0.009} & 0.005      &  & 1.321 & 0.558 & 0.082      &  & 1.233 & 0.483 & 0.192\\
   Ignatius      &  & \textbf{0.001} & 0.010 & \textbf{0.001}      &  & 0.033 & 0.032 & 0.005      & & 0.026 & \textbf{0.005} & 0.002      &  & 0.736 & 0.324 & 0.029      &  & 0.533 & 0.240 & 0.085\\ \hline
   Mean       &  & \textbf{0.020} & \textbf{0.015} & \textbf{0.000}      &  & 0.041 & 0.069 & 0.004      & & 0.080 & 0.038 & 0.006      &  & 1.046 & 0.441 & 0.078      &  & 1.735 & 0.477 & 0.123\\
   % \bottomrule
  \hline
\end{tabularx}
% \captionsetup{type=table, aboveskip=5pt}
\caption{
    \textbf{Pose accuracy on Tanks and Temples}. COLMAP poses are used as ground truth with all methods evaluated using the same protocol. Units: $\text{RPE}_r$ (degrees), ATE (ground truth scale), $\text{RPE}_t$ (scaled by 100). Best results are highlighted in bold.}
\label{table:tanks_pose}
\end{table*}

\subsection{Regularization}
In Figure \ref{fig:densification}, we observe that Gaussians tend to fit high-frequency signals, resulting in a spiky appearance. Similar to \textbf{PhysGaussian} \cite{PhysGaussian}, we apply anisotropy regularization, as described in Equation \ref{eq:spicky}, to control the ratio between the maximum axis $\max(S_i)$ and minimum axis $\min(S_i)$ of the Gaussians, preventing them from becoming spiky. Here, $r$ represents the minimum ratio. Furthermore, to prevent Gaussians from being trapped in local minima due to gradient oscillations, we draw inspiration from \textbf{AbsGS} \cite{AbsGS} and use the absolute gradient $|\frac{\partial L}{\partial \mu^{2D}}|$ to encourage Gaussian splitting, allowing them to search for a better position.
\begin{equation}
    \footnotesize
    \mathcal{L}_{\text{aniso}}=\frac{1}{N}\sum\limits_{i=1}^{N} \max\left\{\frac{\max(S_i)}{\min(S_i)},r\right\}-r
    \label{eq:spicky}
\end{equation}

\section{Experiments}
% rough state what we show in this section
In this section, we %accelerate data preprocessing with \textbf{Structure from Motion (SfM)} and 
compare our approach with existing joint refinement models: \textbf{BARF} \cite{BARF}, \textbf{CF3DGS} \cite{CF3DGS},\textbf{ Nope-NeRF} \cite{NoPe}, and \textbf{SC-NeRF} \cite{SCNeRF}, using \textbf{the Tanks and Temples} \cite{Tanks} and \textbf{CO3DV2} \cite{CO3D} datasets. We also conduct an ablation study to highlight key components of our method. Moreover, we show that our method outperforms \textbf{3DGS} \cite{3DGS}, even when it uses camera pose estimates from \textbf{COLMAP} \cite{COLMAP}, which is often regarded as ground truth to evaluate the effectiveness of pose estimation. Additional experimental results are provided in the supplementary material.

% Table3 CO3D_V2 result
\begin{table*}[t]
    \small
    \centering
    \begin{tabularx}{\textwidth}{l|XXXccc|XXXccc|XXXccc}
    \hline
    \multirow{2}{*}{Scenes} 
    & \multicolumn{6}{c|}{KeyGS (Ours)} 
    & \multicolumn{6}{c|}{Nope-NeRF} 
    & \multicolumn{6}{c}{CF3DGS} \\
    \cline{2-19}
    & PSNR & SSIM & LPIPS & $\text{RPE}_t$ & $\text{RPE}_r$ & ATE 
    & PSNR & SSIM & LPIPS & $\text{RPE}_t$ & $\text{RPE}_r$ & ATE 
    & PSNR & SSIM & LPIPS & $\text{RPE}_t$ & $\text{RPE}_r$ & ATE \\
    \hline
    Apple       
    & \textbf{33.53} & \textbf{0.94} & \textbf{0.07} & \textbf{0.026} & \textbf{0.116} & \textbf{0.001} 
    & 26.86 & 0.73 & 0.47 & 0.400 & 1.966 & 0.046 
    & 29.69 & 0.89 & 0.29 & 0.140 & 0.401 & 0.021 \\
    Bench       
    & \textbf{26.35} & \textbf{0.73} & \textbf{0.30} & \textbf{0.060} & \textbf{0.332} & \textbf{0.002} 
    & 24.78 & 0.64 & 0.55 & 0.326 & 1.919 & 0.054 
    & 26.21 & \textbf{0.73} & 0.32 & 0.110 & 0.424 & 0.014 \\
    Hydrant     
    & \textbf{25.33} & \textbf{0.80} & \textbf{0.15} & \textbf{0.005} & \textbf{0.042} & \textbf{0.000} 
    & 20.41 & 0.46 & 0.58 & 0.387 & 1.312 & 0.049 
    & 22.14 & 0.64 & 0.34 & 0.094 & 0.360 & 0.008 \\
    Skateboard  
    & \textbf{32.74} & \textbf{0.93} & \textbf{0.16} & \textbf{0.029} & \textbf{0.165} & \textbf{0.001} 
    & 25.05 & 0.80 & 0.49 & 0.587 & 1.867 & 0.038 
    & 27.24 & 0.85 & 0.30 & 0.239 & 0.472 & 0.017 \\
    Teddybear   
    & \textbf{32.67} & \textbf{0.93} & \textbf{0.09} & \textbf{0.037} & \textbf{0.120} & \textbf{0.001} 
    & 28.62 & 0.80 & 0.35 & 0.591 & 1.313 & 0.053 
    & 27.75 & 0.86 & 0.20 & 0.505 & 0.211 & 0.009 \\
    \hline
    Average     
    & \textbf{30.12} & \textbf{0.87} & \textbf{0.15} & \textbf{0.031} & \textbf{0.155} & \textbf{0.001} 
    & 25.14 & 0.68 & 0.48 & 0.458 & 0.771 & 1.291 
    & 26.32 & 0.77 & 0.31 & 0.217 & 0.269 & 0.297 \\
    \hline
    Time & \multicolumn{6}{c|}{\textbf{10 min.}} & \multicolumn{6}{c|}{ 30 hr.} & \multicolumn{6}{c}{2 hr.} \\
    \hline
    \end{tabularx}
\caption{
    \textbf{Novel view synthesis and pose accuracy on CO3D V2}. Each baseline method is trained with its public code under the original settings and evaluated with the same evaluation protocol. And COLMAP poses are used as ground truth pose. Best results are highlighted in bold.}
\label{table:co3d_nvs_pose_combined}
\end{table*}
% Figure5
\begin{figure}[h]
    \centering
    \includegraphics[width=0.9\columnwidth]{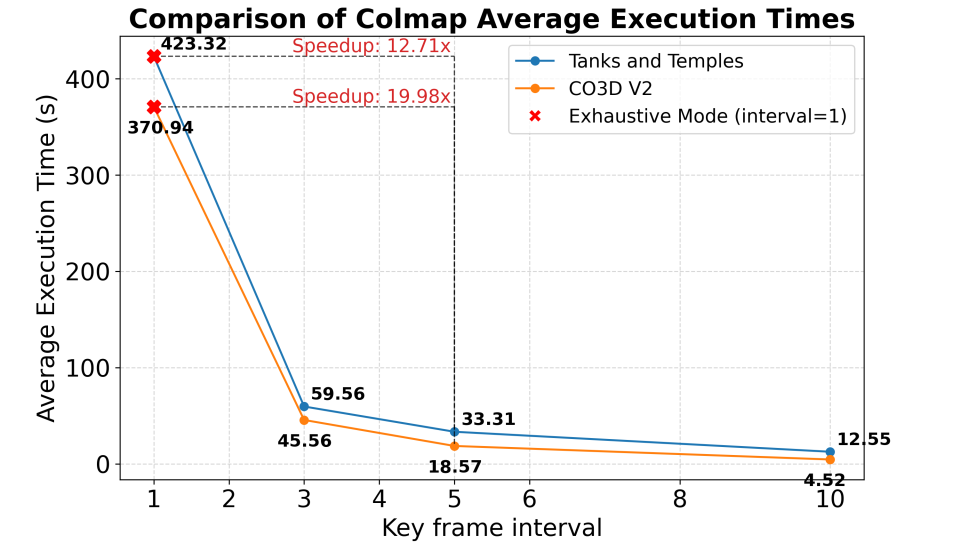}
    % \captionsetup{type=figure, aboveskip=5pt}
    \caption{
        \textbf{KeyFrame-Centric SfM}. Our data preprocessing method can achieve a speedup of at least \textbf{10} times when using the sequential mode compared to the exhaustive mode with full images in \textbf{COLMAP}.}
    \label{fig:keyframe}
\end{figure}

\subsection{Data Preprocessing}
\textbf{COLMAP} is the most commonly used tool for registering camera poses with \textbf{SfM}. The major computational cost arises from bundle adjustment due to the large number of images and feature points, as described in \cite{COLMAP}. We analyze the average computation time for different keyframe subsampling intervals and various downsampling resolutions using all images, as shown in Figure \ref{fig:keyframe}. Both options offer significant speedups compared to the exhaustive mode. However, we found that downsampling image resolution can fail in some scenes because low-resolution images lack robust feature points. Although a keyframe subsampling interval of 10 speeds up the process by over 50 times, it is not robust for outdoor scenes. Therefore, to ensure stability and obtain accurate camera poses, we use a keyframe subsample interval of 5 with full resolution in our experimental setting, it also speeds up the process by 10 to 20 times.

% Figure6
\begin{figure}[h]
    \centering
    \includegraphics[width=0.9\columnwidth]{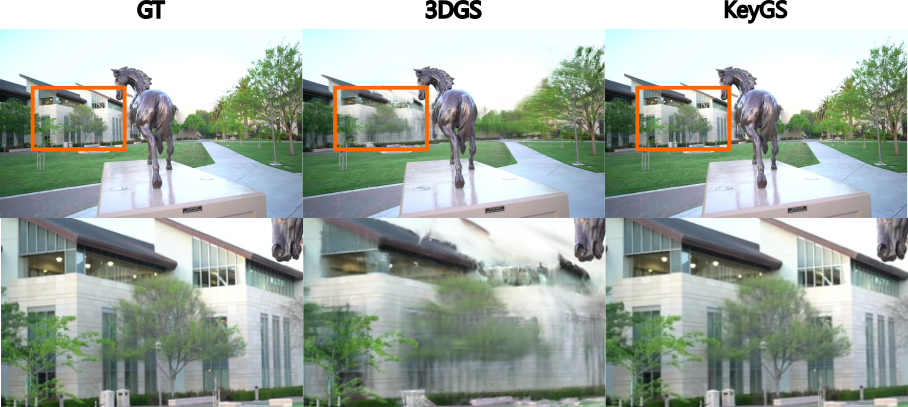}
    % \captionsetup{type=figure, aboveskip=5pt}
    \caption{
        \textbf{3DGS vs. KeyGS.} We show that \textbf{COLMAP} may register noisy camera poses, which can lead to failures in detailed regions, such as the tree in the Horse scene, when using \textbf{3DGS}. In contrast, our method, \textbf{KeyGS}, effectively addresses these issues through joint refinement, even when starting with rough camera poses.}
    \label{fig:3DGS_KeyGS}
\end{figure} 
\subsection{Results}
We use PSNR, SSIM, and LPIPS to evaluate the reconstruction results in our experiments. For camera pose accuracy, we evaluate it with $\text{RPE}_t$ , $\text{RPE}_r$ and $\text{ATE}$. More detailed evaluation metrics are given in the supplementary materials.

For the \textbf{Tanks and Temples} dataset, as shown in Tables \ref{table:tanks_psnr} and \ref{table:tanks_pose}, our method provides very competitive performance and achieves impressively accurate trajectories. Notably, our average \textbf{PSNR} exceeds that of \textbf{CF3DGS} by \textbf{2 dB}.

For the \textbf{CO3DV2} dataset, we followed the experimental setup in \textbf{CF3DGS}, evaluating the same five selected sequences and presenting the results in Tables \ref{table:co3d_nvs_pose_combined}. This dataset is more challenging due to complex trajectories and blurred images. Our method achieves a \textbf{4 dB}higher average \textbf{PSNR} than others and significantly reduces the training time cost. Moreover, our method can continuously refine camera poses, which helps prevent the accumulation of trajectory errors compared to \textbf{CF3DGS}, as illustrated in Figure \ref{fig:teaser}.

\subsection{Ablation Study}
First, we highlight the most significant component of our strategy, as shown in Table \ref{tab:ablation_comp}. The results demonstrate that the \textbf{coarse-to-fine frequency-densification} is the core component of our method. Additionally, both the absolute gradient and anisotropy regularization further improve the performance of our approach.
% compare 3DGS (tanks with gt pose) and KeyGS with key frame 可以show Horse

To demonstrate the importance of joint refinement for camera poses, we compare our method (using a keyframe subsample interval of 5) with \textbf{3DGS} trained using ground truth camera poses from the Tanks and Temples dataset, without joint refinement. The results are shown in Table \ref{tab:ablation_3dgs_keygs}, and an example is depicted in Figure \ref{fig:3DGS_KeyGS}. Our method outperforms the \textbf{3DGS} with the camera poses provided by the Tanks and Temples dataset. This experiment highlights that even when the exhaustive mode in \textbf{COLMAP} is used for pose estimation, inaccuracies in COLMAP-generated camera poses can still degrade \textbf{3DGS} performance.

% Table4, Ablation study of the components of the proposed method 
\begin{table}[h]
  \small
  \centering
  \begin{tabular}{ c  c c c | c c c  }
    % \toprule
    \hline
    & $\ $ \textbf{C2F} $\ $
    & $\ $ \textbf{Abs.} $\ $
    & $\ $ \textbf{Aniso Reg.} $\ $
    & {PSNR$\uparrow$} 
    & {$\text{RPE}_t$$\downarrow$} 
    & {$\text{RPE}_r$$\downarrow$} 
    \\  
    % \midrule
    \hline
     
    (a)
    & \checkmark & \checkmark & \checkmark 
    & \textbf{33.01}
    & \textbf{0.020}
    & \textbf{0.015}
    \\ 
    
    (b)
    &  \checkmark & \checkmark &  
    & 32.94
    & 0.020
    & 0.015
    \\
    
    (c)
    &  \checkmark &   &  \checkmark
    & 31.17
    & 0.020
    & 0.015
    \\
     
    (d)
    &  &  \checkmark &  \checkmark
    & 18.13
    & 0.935
    & 3.022
    \\

    (e) 
    & \checkmark &  &  
    & 31.66
    & 0.020
    & 0.015
    \\

    (f) 
    &  &   & 
    & 19.70
    & 0.534
    & 2.143
    \\

    % \bottomrule
    \hline
  \end{tabular}
  % \captionsetup{type=table, aboveskip=5pt}
  \caption{
    \textbf{Ablation study of the components of the proposed method on the Tanks and Temples dataset.}}
  \label{tab:ablation_comp}
\end{table}
% Table5, Ablation study on COLMAP mode
\begin{table}[h]
    \small
    \centering
        \begin{tabular}{l|ccc}
        % \toprule
        \hline
        Setting & PSNR$\uparrow$ & SSIM$\uparrow$ & LPIPS$\downarrow$\\
        % \midrule
        \hline
        \textbf{3DGS} w/ gt pose from \textbf{COLMAP}& 30.77 & 0.91 & 0.10 \\
        \textbf{KEYGS} w/ rough pose and joint refine & \textbf{33.01} & \textbf{0.94} & \textbf{0.04} \\
        % \bottomrule
        \hline
        \end{tabular}
    % \captionsetup{type=table, aboveskip=5pt}
    \caption{
        \textbf{Ablation study on leveraging exhaustive mode in COLMAP for camera pose estimation and joint refinement within the KeyGS framework.}}
    \label{tab:ablation_3dgs_keygs}
\end{table}

\section{Conclusion}
In this paper, we presented \textbf{KeyGS}, an efficient framework for joint refinement of camera poses and novel view synthesis for monocular image sequences. We analyzed the relationship between \textbf{densification} and \textbf{joint refinement} and proposed the \textbf{coarse-to-fine frequency-aware densification} approach to address gradient oscillation from high-frequency signals. Our approach significantly outperforms previous methods with more accurate novel view synthesis and camera pose estimation as well as drastically reduced training times.

\section*{Acknowledgements}This work was supported in part by the National Science and Technology Council, Taiwan under grants NSTC 111-2221-E-007-106-MY3 and NSTC 113-2634-F-007 002.

\bibliography{aaai25}
\end{document}